\documentclass[letterpaper, 10 pt, conference]{ieeeconf}  

\IEEEoverridecommandlockouts                              

\usepackage{cite}
\usepackage{amsmath,amssymb,amsfonts}
\usepackage{algorithmic}
\usepackage{xcolor}
\usepackage{graphicx}
\usepackage{float}
\usepackage{textcomp}
\usepackage{comment}
\usepackage[normalem]{ulem}
\usepackage{multirow}

\usepackage{subcaption}
\usepackage[dvipsnames]{xcolor}
\usepackage[hidelinks]{hyperref}

\usepackage{url}
\usepackage{hyperref}

\title{\LARGE \bf
From Language to Drives: A Multimodal Framework for Aligning Autonomous Robots with Human Purposes
}

\author{Diana Cerdas-Vargas, Alejandro Romero and Richard J. Duro
\thanks{*This work was funded by the European Union’s Horizon 2020, research and innovation programme under GA 101070381 (`PILLAR-Robots - Purposeful Intrinsically-motivated Lifelong Learning Autonomous Robots'), by Xunta de Galicia (ED431C 2025/32), by the Spanish Science and Education Ministry (PID2024-162223OB-I00), Ministry for Digital Transformation and Civil Service and Next - Generation EU/RRF (TSI-100925-2023-1), and Centro de Investigación de Galicia ``CITIC'' (ED431G 2023/01).}
\thanks{$^{1}$Authors are with the Integrated Group for Engineering Research (GII) at the CITIC research center, Universidade da Coruña,
        {\tt\small diana.cerdas@udc.es, alejandro.romero.montero@udc.es, richard.duro@udc.es}}%
}

\begin{document}

\maketitle
\thispagestyle{empty}
\pagestyle{empty}

\begin{abstract}
This work explores a multimodal approach to the alignment of the behavior of open-ended learning autonomous robots with human purposes  through the use of large language models (LLMs) as cultural enablers. While autonomous robots can generate their own skills and goals through interaction with the environment, they often lack mechanisms to ensure that these remain relevant and aligned to those of human users. The proposed approach makes use of the general and cultural knowledge contained in LLMs and VLMs to establish a multimodal interactive alignment process in which human purposes, expressed in natural language, are transformed into motivational components of a robot architecture in the form of robot missions and drives that guide the behavior of the robot.
The system is structured into three layers for semantic alignment, mission generation, and drive modeling. This design enables a bidirectional interaction loop: human inputs define high-level purposes, robots expand them into executable missions making use of multimodal information relayed by LLMs and VLMs, and feedback maintains alignment with human intent.
Experiments in simulation and with a physical robot demonstrate that the process produces valid missions and behaviors, supporting effective human–agent alignment in interactive scenarios with autonomous robots.

\end{abstract}


\section{Introduction}

A central objective of cognitive and developmental robotics is to construct artificial agents that acquire increasingly sophisticated competencies through prolonged interaction with their environment. Inspired by human cognitive development, developmental robotics emphasizes intrinsically motivated exploration, incremental skill acquisition, and the progressive organization of behavior \cite{BaldassarreMirolli2013Intrinsicallymotivatedlearning,Santucci2013best}. In parallel, the paradigm of Open-Ended Learning (OEL) formalizes the capacity of autonomous systems to continuously generate new goals and expand their behavioral repertoire beyond predefined tasks or datasets \cite{openended}. Together, these perspectives frame autonomy not as the execution of fixed policies, but as an ongoing developmental process.

Within cognitive architectures, such developmental open-endedness is commonly achieved through motivation-based control systems \cite{romero2025hgrail, becerra2021motivational, romero2023perspective}. In these systems, behavior is regulated by internal motivational variables that define \emph{why} the agent acts, allowing goals and skills to emerge as means to regulate internal states. We refer to these domain-independent motivations as \textit{robot purposes}. 

Robot purposes include designer-embedded \textit{needs}, such as homeostatic requirements and intrinsic motivations that drive exploration and competence acquisition \cite{romero2022autonomous,romero2024autonomous}, as well as externally derived purposes, or \textit{missions}, which reflect what human users expect the robot to accomplish during runtime. Both needs and missions are operationalized through \textit{drives} – quantitative variables encoding the urgency of satisfying a given purpose \cite{romero2020motivation}. By acting to regulate its drives, the robot autonomously discovers goals and learns skills that are grounded in the robot's sensorimotor system. Alignment is achieved when the resulting behavior fulfills human expectations \cite{alignmentpaper}.

However, integrating human purposes into a developmental, open-ended system presents a significant challenge. Human intentions are expressed through natural language – ambiguous, context-dependent, and culturally grounded \cite{intention} – whereas motivational architectures require structured, formally defined representations \cite{MotivationasTool}. This discrepancy creates a semantic gap that may lead to misaligned internal purposes, undermining trust and safety \cite{alignmentpaper}. Explainability \cite{ExplainableArtificial} and grounding strategies \cite{groundingspeech} address aspects of this issue, but they do not fully solve the problem of embedding human purposes into the motivational dynamics that drive developmental learning.

Recent advances in Large Language Models (LLMs) and Vision-Language Models (VLMs) provide powerful tools for interpreting linguistic instructions and relating them to perceptual context. Nevertheless, employing such models as end-to-end robot controllers raises concerns regarding explainability, controllability, safety, and the absence of explicit motivational structure \cite{huang2025trust}. From a developmental perspective, delegating control to monolithic policies risks bypassing the very mechanisms – intrinsic motivation, goal discovery, and incremental learning – that underpin autonomous cognitive growth.

We therefore adopt a complementary view: LLMs and VLMs are integrated not as end-to-end control policies, but as high-level semantic and cultural knowledge resources that provide structured priors to a motivation-based cognitive architecture. These multimodal models support the interpretation and clarification of human purposes, constraining the space of plausible missions. The discovery of specific goals and skills, however, remains grounded in the robot’s developmental learning processes and regulated by explicit drives. In this way, large models inform but do not replace the motivational dynamics that sustain autonomous development.

To operationalize this perspective, we propose a multimodal alignment framework composed of three layers: (1) \textit{semantic alignment}, which interprets and refines human purposes through interactive dialogue; (2) \textit{mission generation}, which translates aligned purposes into structured, domain-consistent missions; and (3) \textit{drive modeling}, which transforms missions into quantitative drive functions embedded in the motivational system. An explicit \textit{alignment need}, implemented as an alignment drive, motivates the robot to seek clarification until a coherent internal representation of the intended purpose is achieved.

The contributions of this work are threefold: (i) a cognitively and developmentally grounded framework for embedding human purposes into a motivational architecture supporting open-ended learning; (ii) the integration of specialized LLM- and VLM-based modules as semantic priors within this architecture; and (iii) experimental validation in simulation and on a physical robotic platform, demonstrating effective human–robot alignment while preserving autonomous developmental processes.

By combining multimodal semantic mediation with explicit motivational regulation, this work contributes to the design of developmental autonomous systems that can sustain long-term growth while remaining aligned with human intentions.

\section{Related work}
\label{sec:Relatedwork}

\subsection{Instruction Following and Foundation Models in Robotics}

Bridging the semantic gap between high-level human purposes and executable robot behavior has become a central research topic in embodied AI. Recent approaches increasingly leverage large-scale foundation models to interpret user instructions and generate robot actions. For instance, 
SayCan \cite{doasisay} combines a Large Language Model (LLM) with a predefined library of robot skills. The LLM ranks candidate skills according to semantic relevance, while a value function evaluates feasibility in the current state. This modular strategy enables execution of long-horizon instructions, provided the required primitive skills are available. However, the system remains constrained by a fixed behavioral repertoire and does not generate new goals or expand its skill space.

Other approaches, such as RT-2 \cite{RT2} move toward tighter integration by training a unified vision-language-action model on Internet-scale and robotic data. The model directly maps multimodal inputs to low-level actions, enabling impressive zero-shot generalization. Nevertheless, this end-to-end formulation lacks explicit motivational or goal representations. The resulting policies are difficult to interpret, regulate, or adapt structurally, raising concerns regarding transparency, controllability, and long-term alignment in open-ended environments.

From another point of view, there are systems that address pragmatic intention understanding. One such case is \cite{intention}, which classifies user instructions according to their explicitness and leverages contextual cues to infer appropriate actions. Another is the work presented in \cite{park2024clara} that consideres introducing an estimation of uncertainty in user commands and tries to increase certainty through dialogue. While improving robustness in human–robot interaction, such approaches typically operate within predefined task templates and do not support autonomous goal discovery. 

Finally, shared autonomy methods \cite{sharedautonomy} adapt robot policies to user preferences through interactive feedback. Although effective for collaboration, they primarily refine existing behaviors rather than reorganizing the agent’s motivational structure or enabling the emergence of new purposes.

\subsection{Developmental Robotics and Open-Ended Learning}

Developmental robotics investigates how autonomous agents can acquire progressively complex competencies through processes inspired by human cognitive development \cite{BaldassarreMirolli2013Intrinsicallymotivatedlearning,Santucci2013best}. Rather than relying on externally specified task objectives, developmental systems are driven by intrinsic motivations such as curiosity, competence improvement, or learning progress. These mechanisms support exploratory behaviors, goal babbling, and the autonomous construction of hierarchical skill repertoires over extended interaction with the environment.

Motivation-based cognitive architectures operationalize these principles through internal drives that regulate behavior and trigger self-generated goals \cite{romero2025hgrail, becerra2021motivational, romero2023perspective}. Within the broader Open-Ended Learning (OEL) paradigm \cite{openended}, agents continually expand their behavioral and representational spaces, discovering novel goals and refining predictive models through sensorimotor experience. Crucially, learning remains grounded: goals emerge from embodied interaction rather than from predefined symbolic plans.

Despite these advances, most developmental approaches assume that high-level purposes are either hard-coded by the designer or provided in structured form. While they excel at autonomous goal discovery and skill acquisition \cite{romero2025hgrail, becerra2021motivational, MotivationasTool}, they do not directly address how abstract, linguistically expressed human purposes can be integrated into the agent’s motivational system without disrupting its developmental dynamics. As a result, the problem of semantically aligning human intentions with intrinsically motivated, open-ended learning processes remains largely unresolved.



\subsection{Alignment in Embodied Agents}

Beyond robotics, the AI alignment literature highlights the difficulty of ensuring that artificial agents pursue objectives consistent with human intentions \cite{alignmentpaper}. Many alignment strategies for large models rely on techniques such as reinforcement learning from human feedback \cite{kaufmann2024survey} or instruction tuning \cite{zhang2026instruction}. Although effective at shaping output behavior, these methods do not necessarily yield explicit, inspectable representations of purpose within the agent’s control architecture. In embodied, open-ended systems, alignment requires mechanisms that integrate human purposes into the internal motivational dynamics governing goal generation and learning.

Taken together, existing research either (i) leverages large multimodal models for instruction following, often through fixed skills or end-to-end policies, or (ii) develops developmental architectures capable of autonomous goal discovery without addressing semantic alignment with human purposes. What remains lacking is an integrated framework in which multimodal models provide semantic and cultural priors while an explicit motivational architecture preserves grounded developmental learning, and  alignment.

Our work addresses this gap by embedding LLMs and VLMs as knowledge resources within a motivation-based cognitive architecture that integrates information from different sensing modalities, enabling the transformation of human purposes into aligned drives that guide autonomous open-ended development.

\section{Proposed Approach}

\label{sec:proposal}

\begin{figure*}
\centering
  \includegraphics[width=1\linewidth]{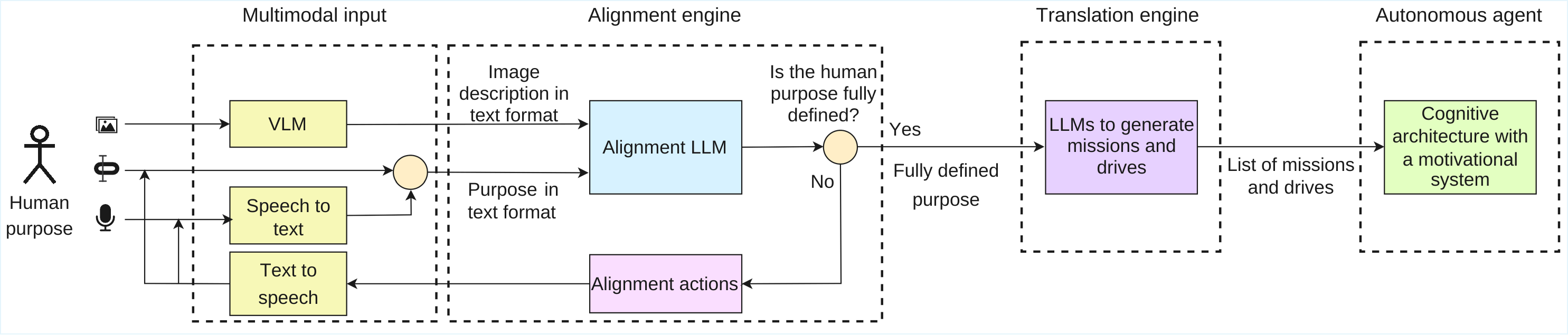}
  \caption{Overview of the proposed framework. The Alignment Engine manages human–robot interaction to align human and robot purposes. The Translation Engine converts aligned purposes into missions and drives, which are integrated into the Motivational System  of a cognitive architecture to guide autonomous robot behavior.}
  \label{fig:LLMdiagram}
\end{figure*}

Translating human purposes expressed in natural language into motivational representations such as missions and drives is not a trivial task. Humans rely on cultural and contextual knowledge to interpret intentions and desires, whereas computational systems lack this implicit background. To bridge this semantic gap, we use LLMs as cultural knowledge bases related to different sensing modalities that, through an alignment process, are able to clarify human purposes.

Fig. \ref{fig:LLMdiagram} illustrates the proposed framework. At the input of the framework there are several elements that allow the user to add multimodal inputs such as an image, language and audio. Next, at its core, the \textit{Alignment Engine} extracts and interprets the human purpose from multimodal information (ususally speech and vision). In cases of ambiguity, it engages in iterative dialogue with the user to refine the interpretation, seeking confirmation before proceeding. Once the purpose is clarified, it is passed to the \textit{Translation Engine}, which decomposes it into missions and corresponding drive functions. These are explicitly instantiated within the motivational system of the cognitive architecture, guiding the robot’s autonomous behavior.

The system is designed for non-expert users. Interaction is therefore concentrated at the specification stage: the user expresses the desired objective in natural language, while the architecture autonomously generates the underlying missions and drives. This allows intuitive purpose specification without requiring knowledge of the robot’s  motivational structure.

The function and configuration of each of the three instances of the LLMs used are explained in detail below. 
Additionally, the prompts used in each of the LLM instances, as well as the source code and the results from the design tests are all available in a public repository 
\footnote{\href{https://github.com/DianaCerdasV/ICDL_From-Language-to-Drives}{GitHub repository}}.

\subsection{Multimodal Input} 

This module manages multimodal inputs to improve the user experience by providing a more intuitive interaction interface. Visual data is processed through a Vision Language Model (VLM), specifically we used \textit{GPT-4o} by OpenAI. This model is configured through a specialized prompting to generate a detailed scene description and provide an inferred human intent. For audio processing, the system integrates two distinct frameworks: OpenAI's \textit{Whisper} model for Speech-to-Text (STT) transcription, and \textit{Google Text-to-Speech} (gTTS) to handle the system’s vocal responses.

\subsection{Alignment Engine}

This first LLM uses an instance of the \textit{phi4:14b} model by Ollama. It receives the natural language purpose provided by the user and/or a visual description of the environment and reformulates it into a clear, unambiguous specification. When necessary, it engages in a brief interactive dialogue with the user to resolve ambiguities or request missing details. 

\subsubsection{Prompt}

The prompt used to guide this LLM instance was carefully structured to include all relevant contextual and operational information. It consists of six main components: (1) Context, which introduces the LLM's task and defines key concepts such as needs, missions and drives; (2) Robot description and general description of its environment; (3) Robot perceptions, specifying available sensory inputs along with their value ranges; (4) Input description, explaining the nature and format of the user-provided purpose; (5) Task constraints, which define what the LLM must do and explicitly limit its scope; and (6) Interaction rules, which describe the expected structure and format of the model’s responses, ensuring consistency and compatibility with downstream components.

\subsubsection{Input and Output Configuration}
The input to this module is a purpose expressed in natural language by the user, along with the image description from the Multimodal Input, when used. The user is required to answer questions and provide additional details to the LLM when needed. 

To determine when a purpose was fully aligned 
we looked at how humans clarify intentions during conversation. Often, they ask: \textit{“Is this what you meant?”} or \textit{“Is this what you want?”}. Based on this behavior, the user is who decides when the purpose is correctly understood. From a human-robot interaction perspective, this means the human user has full control over when to move to the next stage.

The output is a final aligned and refined purpose statement, written as a clear paragraph. If, based on the robot’s capabilities and perceptions, the model determines that the purpose is not achievable, it returns a standard message: \textit{``It is impossible to achieve your purpose with this robot''.}

\subsection{Mission Generator}

Once the user purpose has been aligned, a second LLM instance, also using the \textit{phi4:14b} model, is responsible for decomposing it into a structured set of missions. These missions represent concrete end-states that the robot must achieve in order to fulfill the user’s intention.

Each mission also includes an associated weight that indicates its relative importance within the overall task context, reflecting how critical it is for achieving the human’s intended purpose for the robot. The output of this component serves as a bridge between high-level human intent and the motivational structure that will later govern the robot’s autonomous behavior.

\subsubsection{Prompt}

The prompt for this instance was designed to clearly instruct the model to generate a list of missions that directly correspond to the aligned purpose. It included: (1) Context, which introduces the LLM's task and defines key concepts such as needs, missions and drives; (2) Robot description; (3) Robot perceptions; (4) Requirements for creating missions and assigning their weight (priority to fulfill the purpose); (5) Interaction rules and output format; and (6) A list of predefined internal needs, introduced by the designer within the cognitive architecture.

\subsubsection{Input and Output Configuration}

The input for this module is the refined purpose generated by the Alignment Engine, expressed as text. On the other hand, the output is an array of missions, where each mission is represented as another array. The first element is a descriptive tag of the mission, it should represent a meaningful end-state, for example \textit{``object\_in\_hand"}.  The second element corresponds to a numerical weight that indicates the relative priority of that mission. The format of the output is: \textit{$[[tag_1, weight_1], [tag_2, weight_2], ..., [tag_N, weight_N]]$}.

\subsection{Drive Generator}

A third LLM instance uses the \textit{GPT-4.1} model from OpenAI, and is responsible for modeling a mathematical drive function for each mission. A drive function is a continuous, monotonic mathematical expression that quantifies the urgency to satisfy a given mission based on the robot’s current state. This function must return zero only when the mission has been fulfilled, and increase smoothly as the robot deviates from the desired goal-state. The output of this component integrates directly into the robot’s motivational system.

\subsubsection{Prompt}

The prompt provided to this LLM was designed to ensure internal consistency with the elements of the cognitive architecture, and alignment with the robot’s perceptual space. It includes: (1) Context definitions for missions, drive functions, and drive values; (2) Robot description; (3) Robot perceptions; (4) Format specification of the missions generated in the previous stage; and (5) Structured instructions for generating a drive function for each mission.

\subsubsection{Input and Output Configuration}

The input to this component consists of two elements. First, the fully refined purpose provided by the Alignment Engine, expressed in text form. Second, the list of missions generated by the previous LLM instance.
The output is a list of mathematical drive functions, one for each mission. Each entry follows the structure: \texttt{Mission: [tag, weight]} and \texttt{Drive: valid Python expression}. If the LLM detects that the purpose cannot be fulfilled with the robot’s current capabilities, it returns the message: \textit{``It is impossible to achieve your purpose with this robot''}.

\section{Experiments and Results}
\label{sec:experiments}

To evaluate the proposed framework, we conducted two experiments. The first experiment evaluates whether the framework can align with and correctly interpret human purposes, while the second assesses the generation of missions and drives and their application to robotic tasks. 

\subsection{Dataset}
\label{sec:dataset}

To evaluate whether the proposed framework can align with and interpret human purposes expressed with different levels of clarity, we defined three robot-executable purposes with increasing complexity. These purposes were: arranging apples by color (simple), classifying cylinders by color into baskets (intermediate), and stacking cylinders at the center of the table from largest to smallest (complex).


To simulate how a non-expert user might express these purposes in natural language, we created 15 different ambiguous user inputs for each of the three purposes. Table \ref{tab:ambiguous_inputs} presents example ambiguous inputs for each purpose. In addition, three different images illustrating the intended outcome were used for each human purpose. During the experiments, the images associated with each purpose were alternated across trials so that the VLM was exposed to different visual scenarios.

\begin{table}[htb]
\centering
\caption{Example ambiguous user inputs for the three evaluated purposes, simulating how non-expert users describe tasks in natural language.}
\begin{tabular}{|p{3.4cm}|p{4,2cm}|}
\hline
\textbf{Purpose} & \textbf{Example of non-expert user input} \\ \hline
Arrange apples by color & \textit{``Place the apples where they belong [as in the image]"} \\ \hline
Classifying cylinders by color & \textit{``Sort the cylinders on the table [as in the image]."} \\ \hline
Stacking cylinders in the center of a table from biggest to smallest & \textit{``Arrange the pieces on top of each other [as in the image]"} \\ \hline
\end{tabular}
\label{tab:ambiguous_inputs}
\end{table}

The complete dataset, including the 45 expressions and 9 images used in the experiments, is available in a public repository \footnote{\href{https://github.com/DianaCerdasV/ICDL_From-Language-to-Drives}{GitHub repository}}.

\subsection{Semantic Alignment Test}
\label{sec:align_test}

This test was designed to evaluate the  ability of the Alignment Engine to disambiguate natural language purposes and convert them into fully defined, structured task descriptions through user interaction. Since both the inputs and outputs of this process are purely linguistic, 
cosine semantic similarity in embedding space \cite{Wang2020} was used to measure how closely their meanings align.  

Cosine semantic similarity 
captures the similarity of two sentences by comparing the angle between their embedding vectors.
A value of 1 indicates identical semantic meaning, a value of 0 indicates no semantic relation, and a value of -1 indicates opposite meanings.




To evaluate the alignment process, we tested the Alignment Engine using the dataset described in Section \ref{sec:dataset}. Each input was processed under four distinct experimental conditions to isolate the impact of multimodal input and human feedback:

\begin{enumerate}
\item \textbf{Full Interaction (Multimodal + Human Feedback)}: The system uses multimodal input, such as an image illustrating the intended outcome, together with active user interaction, where the LLM performs iterative clarification until the user confirms that the purpose is fully understood.
\item \textbf{Autonomous Multimodal Interaction (Multimodal + No Human Feedback)}: The LLM interprets the input together with the multimodal information as provided, without any subsequent human clarification.
\item \textbf{Human Interaction (Unimodal + Human Feedback)}: The system operates without multimodal support but maintains iterative dialogue with the user to refine and disambiguate the purpose.
\item \textbf{Baseline (Unimodal + No Human Feedback)}: The LLM interprets the ambiguous textual purpose without additional resources or user feedback.
\end{enumerate}

In each of the four experimental conditions, the final purpose description generated by the LLM was compared with an ideal manually written reference description using cosine semantic similarity. The embedding vectors were computed using the \textit{all-MiniLM-L6-v2} model \cite{sbert2023}.
The reference purpose descriptions used for semantic comparison are listed in Table \ref{tab:task_prompts}.

\begin{table}[H]
\centering
\caption{Detailed ideal purpose descriptions used as reference to estimate the semantic similarity.}
\label{tab:task_prompts}
\renewcommand{\arraystretch}{1.5}
\begin{tabular}{|p{7.5cm}|}
\hline

Arrange apples by color \\
\textit{``The robot should arrange the apples on the table by their color. First, the robot has to grasp the red apple (object2) and place in in the right side of the table ($x\_position>0.5$). After that, the robot has to grasp the green apple (object3) and place it on the left side of the table ($x\_position<0.5$).''} \\
\hline

Classifying cylinders by color into baskets \\
\textit{``The robot must arrange the cylinders by color into baskets. First it has to reach the red cylinder (object2) with its hand, grasp it and pick it up. Then, the robot should move its hand, with the red cylinder picked, to the green basket (object4) and release it into the basket. After that, the robot should move its hand to the blue cylinder and grasp it. Then, the robot should take the blue cylinder (object3) to the position of the yellow basket (object5) and place the blue cylinder into the basket.''} \\
\hline

Stacking cylinders in the center of the table from largest to smallest \\
\textit{``The robot's task is to stack the cylinders in the center of the table by their size (based on their diameters). The largest cylinder goes at the bottom and the smallest one at the top. To do this, first, the robot should grasp the object4 (largest cylinder), move it to the center of the table. After that, the robot should grasp the object2 (second largest cylinder) and place it on top of the object4. Last, it has to grasp the object3 (smallest cylinder), move it to the center of the table and stack it on top of the object2.''} \\
\hline

\end{tabular}
\end{table}

\begin{figure}[ht]
\centering
\includegraphics[width=0.48\textwidth]{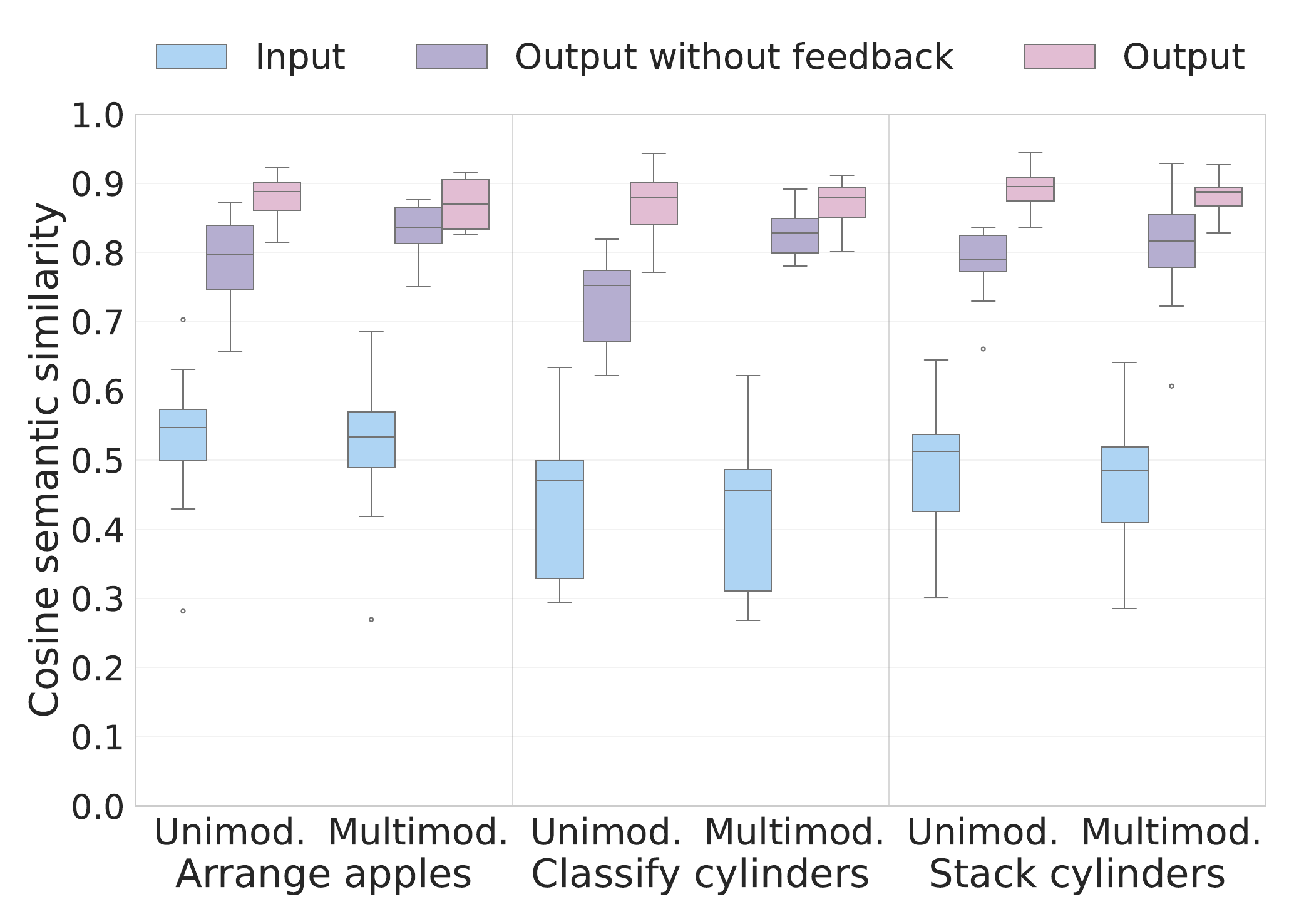}
\caption{Box plot of semantic similarity between input purposes and output descriptions generated by the Alignment Engine under multimodal and unimodal configurations, with and without human feedback.}
\label{fig:boxplot}
\end{figure}

The results of the semantic alignment evaluation are summarized in Fig.\ref{fig:boxplot},  
which illustrates the distribution of cosine similarity scores across the different experimental conditions. For the initial low-detail inputs provided by users, similarity scores ranged between 0.4 and 0.6. These values quantify the initial semantic gap, reflecting the fact that human instructions are often underspecified or ambiguous. At this level of similarity, the Alignment Engine cannot reliably translate purposes into well-defined missions and drive functions.

When the purpose was interpreted by the LLM without alignment interaction, the cases that included multimodal input showed much higher semantic similarity to the ideal reference. This improvement suggests that the model was able to use visual context about the robot and its environment to enrich the linguistic description. However, despite this increase, the results show that the generated purposes still diverge semantically from the ideal reference. In other words, the improvement is insufficient to guarantee that the generated purpose fully captures the user’s intended meaning.

A Mann-Whitney U test was performed to determine whether the use of multimodal resources led to greater improvement in semantic alignment. This non-parametric test was used to evaluate distributional differences between the unimodal and multimodal settings for three cases: the initial inputs, the outputs without feedback, and the outputs with feedback. Statistical differences were observed in the no-human-feedback scenarios for the \textit{Arrange apples by color} and \textit{Classify cylinders by color} purposes. In these cases, the p-values were below the \(\alpha = 0.05\) threshold, leading to rejection of the null hypothesis. This result indicates that multimodal resources provide a significant improvement mainly in scenarios without human feedback.

When interactive alignment is enabled (pink distributions in Fig. \ref{fig:boxplot}), cosine similarity scores consistently converge toward high and stable values across all three tasks. Importantly, these scores are significantly higher than those obtained in non-interactive conditions, regardless of whether multimodal resources are available. This result highlights the critical role of user interaction in achieving precise semantic alignment. Iterative clarification ensures that the internal representation of the purpose closely matches the intended meaning.

The contribution of multimodal information (e.g., visual input) is more nuanced. In interactive settings, multimodal resources do not substantially increase the final semantic similarity compared to linguistic-only interaction. Instead, their primary benefit lies in improving alignment efficiency. Specifically, multimodal scenarios required an average of 3 interaction rounds to reach convergence, whereas text-only interaction required 4.5 rounds on average, with some cases demanding up to 11 rounds. Thus, multimodal grounding reduces the interaction overhead needed to achieve alignment.

Overall, these results indicate that multimodal information enhances interpretation quality and reduces clarification effort, but interactive alignment remains essential to ensure reliable semantic grounding before translating purposes into motivational structures.

\subsection{Mission and Drive Generation Test}

This test evaluates the Translation Engine’s capacity to transform refined human purposes into structured motivational representations through two interconnected stages: (1) decomposing purposes into missions, and (2) generating mathematical drive functions for each mission. To isolate these stages from spurious effects introduced by the Alignment Engine, no feedback from the mission or drive evaluation was used to further disambiguate the purpose during these tests.

The inputs to the mission generator were the aligned purposes obtained from the experiment described in Section \ref{sec:align_test}. The resulting list of missions was then used as input to the drive generation stage, allowing this second step to be evaluated independently from unrelated input errors. 
To assess the validity of the missions and drives, an expert manually classified each output as valid or invalid. To this end, several factors were considered. First, we assessed whether the complete set of missions would enable the robot to achieve the intended human purpose and whether the associated weight distribution was appropriate. Additionally, each generated drive function was individually examined to ensure both mathematical correctness and proper implementation in Python. It was also verified that each function reached zero, or decreased, only under the appropriate task conditions. For clarity, Table \ref{tab:ideal_missions} presents the set of missions and drives obtained for the ideal \textit{Arrange apples by color} purpose defined in Table~\ref{tab:task_prompts}.

\begin{table}[htb]
\centering
\footnotesize
\caption{Missions and drive functions for the \textit{Arrange apples by color} ideal purpose.}
\label{tab:ideal_missions}
\begin{tabular}{|c|c|}
\hline
\textbf{Mission [label, weight]} & \textbf{Drive} \\
\hline
["red\_apple\_in\_hand", 0.2] & 
\begin{tabular}[c]{@{}l@{}}
$|gripper - 1| + |x_{rob} - x_{red}|$ \\
$+ |y_{rob} - y_{red}|$
\end{tabular} \\
\hline
["red\_apple\_right\_side", 0.4] & 
\begin{tabular}[c]{@{}l@{}}
$\max(0,\, 0.5 - x_{red})$
\end{tabular} \\
\hline
["green\_apple\_in\_hand", 0.6] & 
\begin{tabular}[c]{@{}l@{}}
$|gripper - 1| + |x_{rob} - x_{green}|$ \\
$+ |y_{rob} - y_{green}|$
\end{tabular} \\
\hline
["green\_apple\_left\_side", 0.8] & 
\begin{tabular}[c]{@{}l@{}}
$\max(0,\, x_{green} - 0.5)$
\end{tabular} \\
\hline
\end{tabular}
\end{table}


The first two missions in Table \ref{tab:ideal_missions} are related to grasping and placing the red apple in the corresponding location, while the last two focus on grasping and placing the green apple in its correct position. Overall, the weight distribution follows an increasing pattern, consistent with the training prompt, where later missions are assigned higher weights to reflect the intended task order. For readability, the drive functions are presented in simplified form, although they were originally expressed in Python code.

It can be observed that the drive functions associated with grasping evaluate to zero only when the robot is holding an object (\(gripper = 1\)) and the \((x, y)\) positions of the apple and the robot hand coincide. On the other hand, the drives associated with placing the apples are structurally similar. For the red apple, the function depends on its \(x\)-position: if \(x_{red} < 0.5\) (left side of the table), the drive is non-zero, and it becomes zero only when \(x_{red} > 0.5\), thus ensuring placement on the right side of the table. A similar behavior is observed for the green apple, but in this case the goal is the opposite: the drive encourages the apple to be placed on the left side of the table \((x_{green} < 0.5)\).

\begin{figure}[htb]
    \centering
    \includegraphics[width=\linewidth]{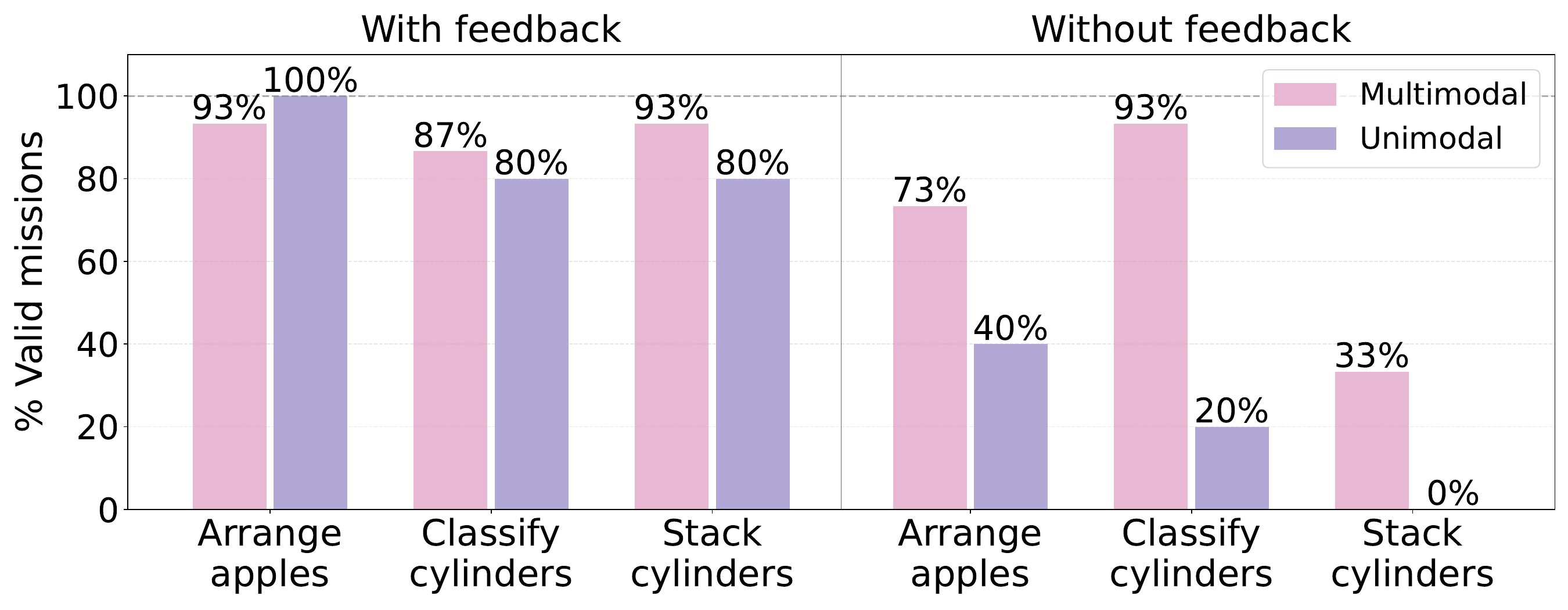}
    \caption{Validity of generated missions across the three tested purposes. The ideal purpose (Table \ref{tab:task_prompts}) had a validity rate of 100\% across the three purposes.}
    \label{fig:missions_validity}
\end{figure}



\begin{figure}[htb]
    \centering
    \includegraphics[width=\linewidth]{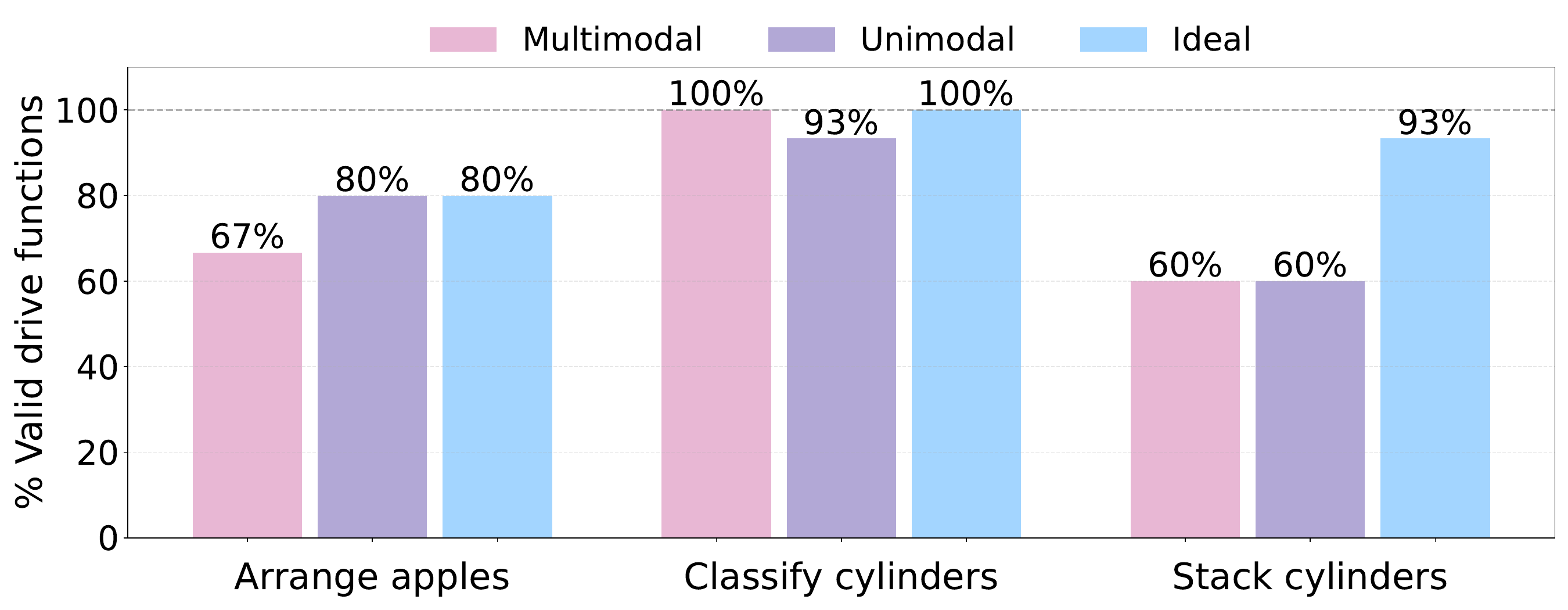}
    \caption{Validity of generated drive functions across the three tested purposes and the success rate for the ideal purposes (Table \ref{tab:task_prompts}).}
    \label{fig:drives_validity}
\end{figure}

A similar manual analysis was conducted for each set of missions and drives. 
As illustrated in Fig.\ref{fig:missions_validity}, scenarios involving human feedback led to a significantly higher percentage of valid missions than scenarios without user intervention. Although the autonomous multimodal configuration (multimodal without feedback) occasionally exhibited superior performance, its success remains highly dependent on the visual context provided and the complexity of the task. Consequently, this setting cannot guarantee consistent reliability. Therefore, only scenarios involving human feedback were used for the final drive evaluation.
Regarding the generation of drive functions, Fig.~\ref{fig:drives_validity}shows that validity rates varied across tasks. While the system performed very well on the \textit{Classify cylinders by color} task, it was less successful on the \textit{Stack cylinders by size} task, failing in 6 out of 15 instances. Compared with the ideal case, performance on the first two purposes closely matches the ideal results, whereas performance on the stacking task is substantially worse in both the multimodal and unimodal settings. The reason for this behavior is that, as the purpose becomes more complex and requires a more detailed description, the Alignment Engine becomes more prone to producing ambiguous purposes. When these ambiguities propagate to the drive generation stage of the Translation Engine, they lead to inconsistent performance and, in some cases, invalid drives.

We observed that when the purpose is well specified, repeated runs through the Translation Engine produce essentially the same valid set of drives. However, if the purpose contains ambiguity, repeated runs through the Translation Engine often produce different sets of drives, some of which are invalid. This variability can therefore be interpreted as a signal that the purpose remains underspecified, which could be used to trigger further dialogue between the Alignment Engine and the user.


In general, these findings demonstrate that the Translation Engine can adequately translate fully defined user purposes into motivational structures. When failures occur, they are usually caused by ambiguities in the purpose definition, which indicates that the Alignment Engine should be triggered to refine that definition.

\subsection{Autonomous Robot Test}

Once the complete framework, including the three LLM instances for alignment, mission generation, and drive modeling, had been validated, it was tested on a robot to evaluate the full system operation. The framework was integrated into the e-MDB cognitive architecture \cite{becerra2021motivational, MotivationasTool}, which has a motivational system based on needs, missions, and drives.

\begin{figure}
\centering
\begin{minipage}{.5\linewidth}
\centering
{\includegraphics[width=\linewidth]{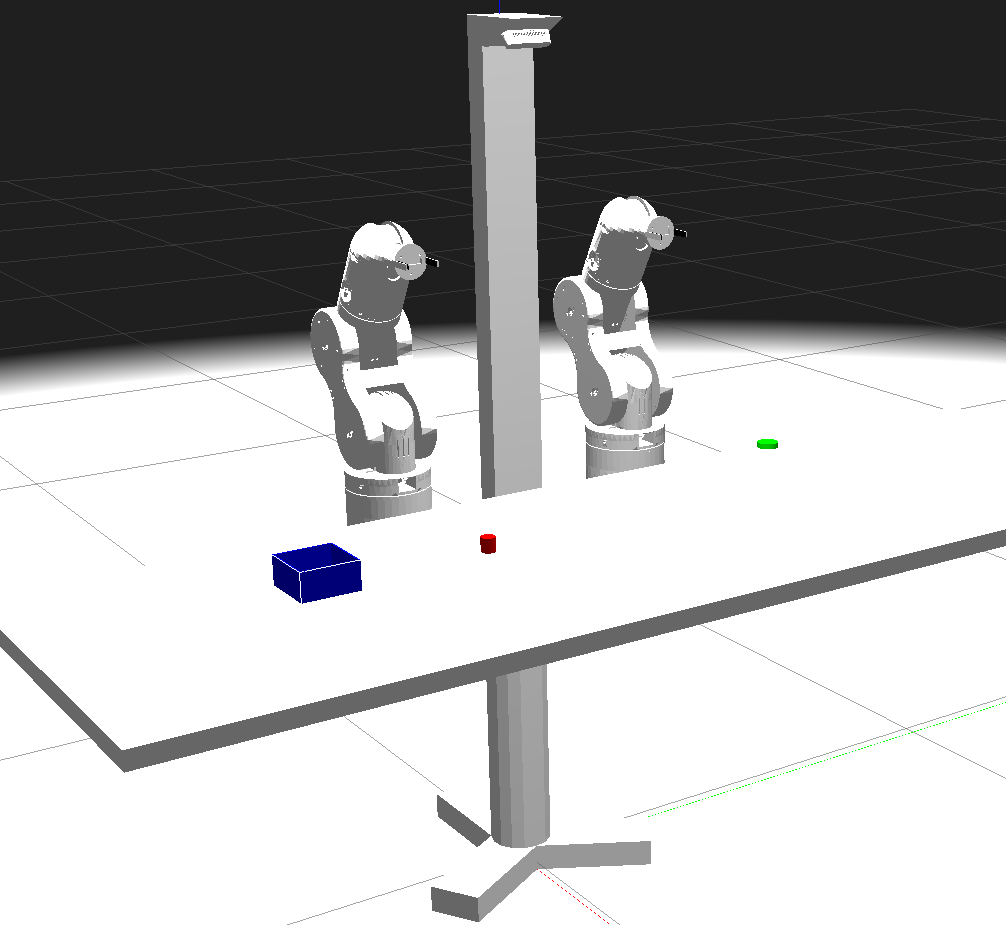}}
\end{minipage}%
\begin{minipage}{.495\linewidth}
\centering
{\includegraphics[width=\linewidth]{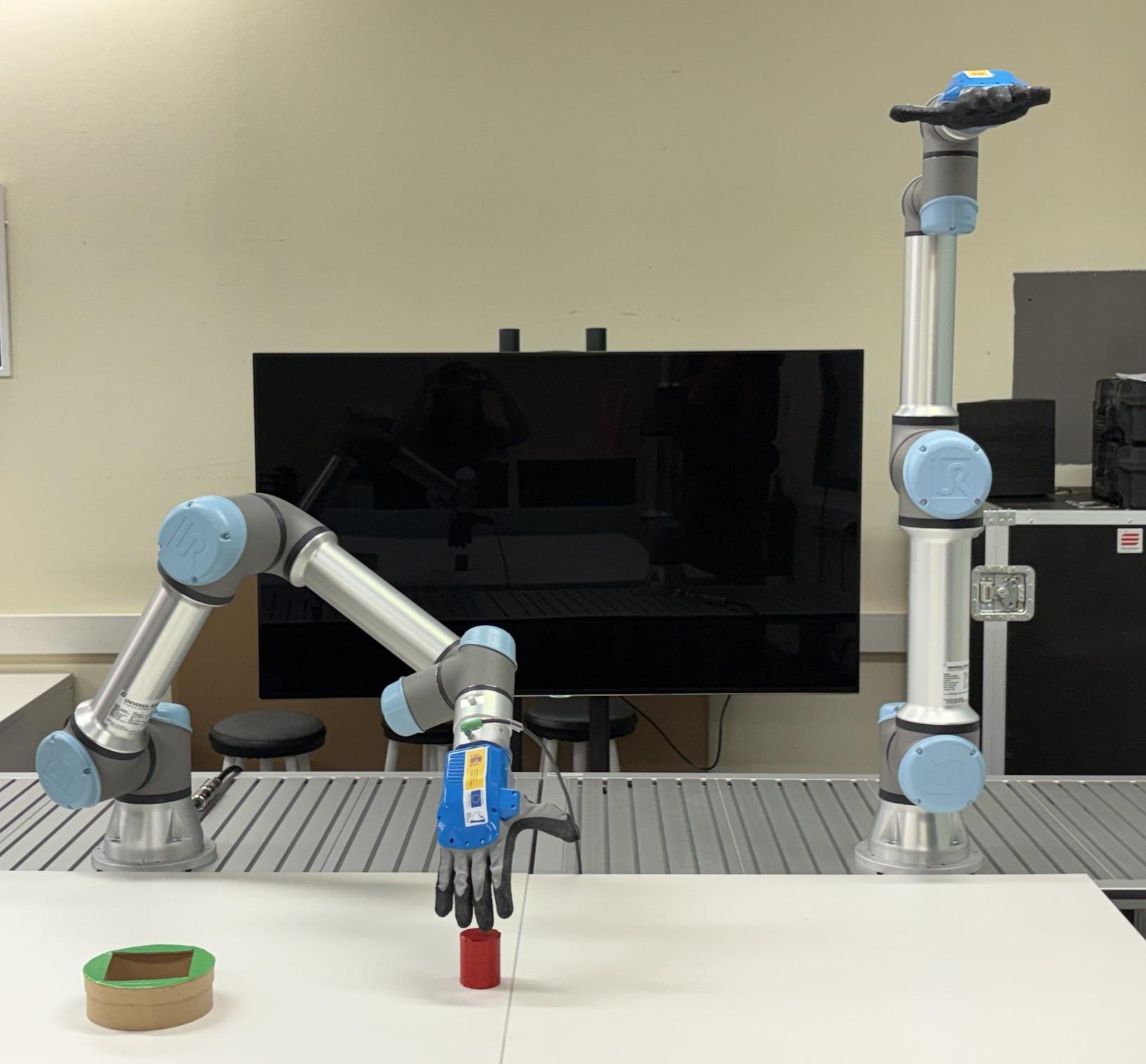}}
\end{minipage}%
\caption{Simulated and real environment used to validate the integration of the generated missions and drives within the e-MDB cognitive architecture.}
\label{fig:sim}
\end{figure}

In this experiment, the robot’s purpose was \textit{to place a red cylinder inside a box and press a green button whenever the red cylinder was outside its reachable area}. The system interpreted the human purpose, generated the missions and drive functions, and executed the task autonomously. Fig. \ref{fig:sim} shows the experimental setup. A video of the robot’s performance is available in a public repository \footnote{\href{https://github.com/DianaCerdasV/ICDL_From-Language-to-Drives/tree/main/Videos}{Videos}}.


\begin{figure}[ht]
  \centering
  \includegraphics[width=0.45\textwidth]{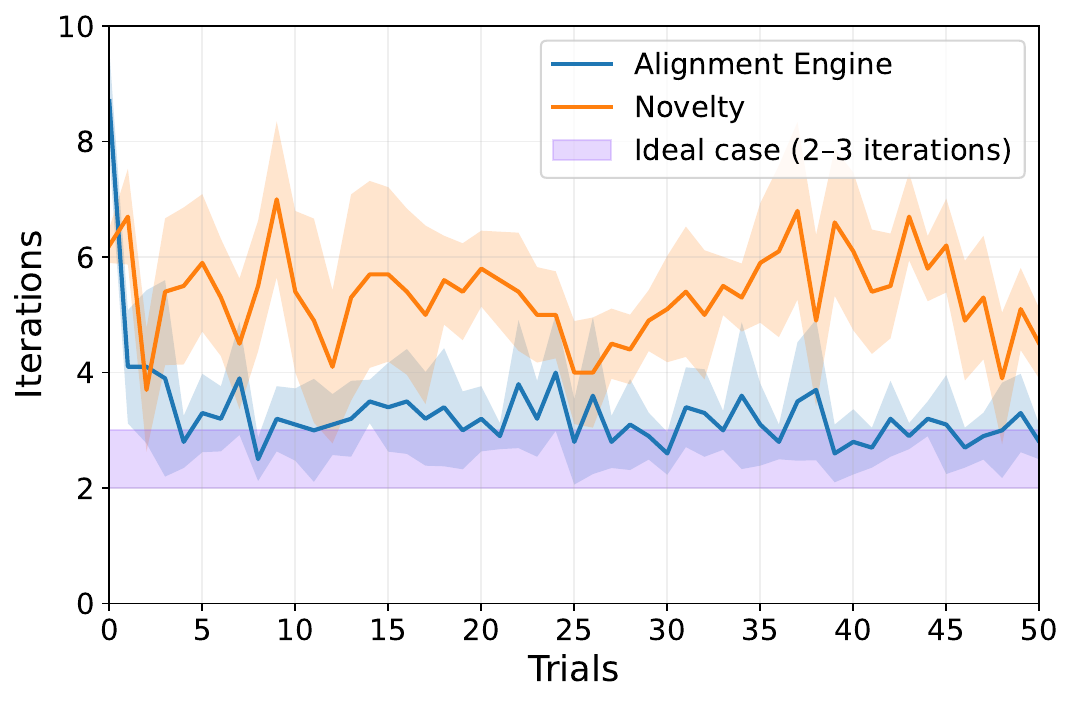}
  \caption{Number of iterations required for the robot to achieve the assigned purpose using the Alignmnet and Translation Engines compared to using just novelty.} 
  \label{fig:trials}
\end{figure}

The experiment was run ten times to check the consistency of the robot’s behavior. To obtain a performance metric, each run of the experiment included 50 trials, in which we recorded how many iterations the robot needed to achieve the human purpose in each case. Fig.\ref{fig:trials} shows these results.

The results shown in Fig.~\ref{fig:trials} indicate that it consistently reached the goal in 3–4 iterations, which is close to the practical minimum given the initial positions of the objects on the table. 
Notably, the robot reached the goal with very few iterations, suggesting a more directed behavior than what would typically be expected from novelty-based exploration alone. Moreover, the robot quickly converges to the correct sequence of actions, guided by the rewards induced by the generated drives. These results demonstrate that the generated missions and drives enabled the robot to complete the task accurately and efficiently.

% ...
 
\section{Conclusions and Future Work}
\label{sec:conclusions}

This paper introduced a multimodal framework for aligning human purposes with the motivational system of an autonomous robot. By integrating three LLM-based components for semantic alignment, mission generation, and drive modeling, the approach enables non-expert users to specify goals in natural language while grounding them in a structured, motivation-driven control architecture.

Experimental results demonstrate that interactive alignment significantly improves semantic precision, confirming that clarification dialogue is essential to reduce ambiguity and reliably translate human purposes into executable motivational structures. While multimodal inputs enhance interpretation efficiency and provide contextual information, iterative human feedback remains crucial for robust alignment. The generated missions and drive functions were found to satisfy the required formal constraints and, when embedded in a cognitive architecture, support consistent goal-directed behavior.

Overall, the proposed system shows the potential of large multimodal models to serve as semantic mediators within a developmental, motivation-based architecture, while maintaining interpretability and alignment. Future work will expand multimodal interaction capabilities and investigate adaptive alignment mechanisms that evolve through prolonged human–robot interaction.

\addtolength{\textheight}{-12cm}   





\bibliographystyle{IEEEtran}
\bibliography{References_ICDL2026}

\end{document}